\pdfoutput=1

\documentclass[11pt]{article}

\usepackage[]{emnlp2021}

\usepackage{times}
\usepackage{latexsym}

\usepackage[T1]{fontenc}

\usepackage[utf8]{inputenc}

\usepackage{microtype}

\usepackage{graphicx}
\usepackage{float}
\usepackage{subfigure}
\usepackage{amsmath}
\usepackage{amssymb}
\usepackage{color}
\usepackage{booktabs} 

\newcommand{\cparagraph}[1]{\vspace{1.5mm}\noindent\textbf{#1}}
\newcommand*\samethanks[1][\value{footnote}]{\footnotemark[#1]}

\title{Learning Neural Templates for Recommender Dialogue System}

\author{Zujie Liang$^{1}$\thanks{\ \ Equal contribution. Work performed when Zujie Liang was an intern at Microsoft STCA.}
{},
Huang Hu$^2$\samethanks, Can Xu$^2$, Jian Miao$^{2}$, Yingying He$^{2}$, 
	\\\textbf{Yining Chen$^2$, Xiubo Geng$^2$, Fan Liang$^1$, Daxin Jiang$^2$\thanks{\ \ Corresponding author: djiang@microsoft.com.}}\\
	$^1$Sun Yat-sen University, Guangzhou, China\\
	$^2$Microsoft Corporation, Beijing, China \\
	{ \tt $^1$\{liangzj9@mail2.sysu.edu.cn, isslf@mail.sysu.edu.cn\}} \\
	{ \tt $^2$\{huahu,caxu,jianm,yingyhe,yinichen,xigeng,djiang\}@microsoft.com} \\}
\date{}

\begin{document}
\maketitle

\begin{abstract}

Though recent end-to-end neural models have shown the promising progress on Conversational Recommender System (CRS), two key challenges still remain. 
First, the recommended items cannot be always incorporated into the generated replies precisely and appropriately. 
Second, only the items mentioned in the training corpus have a chance to be recommended in the conversation. 
To tackle these challenges, we introduce a novel framework called \textbf{NTRD} for recommender dialogue system that decouples the dialogue generation from the item recommendation.
\textbf{NTRD} has two key components, \textit{i.e.}, response template generator and item selector.
The former adopts an encoder-decoder model to generate a response template with slot locations tied to target items,
while the latter fills in slot locations with the proper items using a sufficient attention mechanism.
Our approach combines the strengths of both classical slot filling approaches (that are generally controllable) and modern neural NLG approaches (that are generally more natural and accurate). 
Extensive experiments on the benchmark \textsc{ReDial} show our NTRD significantly outperforms the previous state-of-the-art methods.
Besides, our approach has the unique advantage to produce novel items that do not appear in the training set of dialogue corpus. The code is available at \url{https://github.com/jokieleung/NTRD}.

\end{abstract}
\section{Introduction}

Building an intelligent dialogue system that can freely converse with human, and fulfill complex tasks like movie recommendation, travel planning and etc, has been one of longest standing goals of natural language processing (NLP) and artificial intelligence (AI). 
Thanks to the breakthrough in deep learning, the progress on dialogue system has been greatly advanced and brought into a new frontier over the past few years. 
Nowadays, we are witnessing the booming of virtual assistants with conversational user interface like Microsoft Cortana, Apple Siri, Amazon Alexa and Google Assistant. 
The recent large-scale dialogue models such as DialoGPT \cite{zhang2019dialogpt}, Meena \cite{adiwardana2020towards} and Blender \cite{roller2020recipes}, demonstrate the impressive performance in practice.
Besides, the social bots such as XiaoIce~\cite{shum2018eliza} and PersonaChat~\cite{zhang2018personalizing} also exhibit the great potential on the emotional companion to humans. 

The conversational techniques shed a new light on the search and recommender system, as the users can seek information through interactive dialogues with the system.
Traditional recommender systems often rely on matrix factorization methods \cite{koren2009matrix,rendle2010factorization,wang2015collaborative,he2017neural}, and suffer from the cold-start problem \cite{schein2002methods,lika2014facing} when no prior knowledge about users is available.
On the other hand, existing recommendation models are trained on offline historical data and have the inherent limitation in capturing online user behaviors~\cite{yisong2020advanced}.
However, the user preference is dynamic and often change with time.
For instance, a user who usually prefers science fiction movies but is in the mood for comedies, would likely get a failed recommendation. 

In recent years, there is an emerging trend towards building the recommender dialogue system, \textit{i.e.}, Conversational Recommendation System (CRS), which aims to recommend precise items to users through natural conversations.
Existing works~\citep{li2018towards,chen2019towards,zhou2020improving,ma2020bridging} on this line usually consist of two major components, namely a recommender module and a dialogue module. 
The recommender module aims at retrieving a subset of items that meet the user’s interest from the item pool by conversation history,
while the dialogue module generates free-form natural responses to proactively seek user preference, chat with users, and provide the recommendation.
To incorporate the recommended items into the responses, a switching network~\citep{gulcehre2016pointing} or copy mechanism~\citep{gu2016incorporating} is utilized by these methods to control whether to generate an ordinal word or an item at each time step.
Such integration strategies cannot always incorporate the recommended items into generated replies precisely and appropriately. 
Besides, current approaches do not consider the generalization ability of the model.
Hence, only the items mentioned in the training corpus have a chance of being recommended in the conversation. 

In this paper, we propose to learn \textbf{N}eural \textbf{T}emplates for \textbf{R}ecommender \textbf{D}ialogue system, \textit{i.e.}, \textbf{NTRD}. NTRD is a neural approach that firstly generates a response ``template'' with slot locations explicitly tied to the recommended items. 
These slots are then filled in with proper items by an item selector, which fully fuses the information from dialogue context, generated template and candidate items via the sufficient multi-head self-attention layers. 
The entire architecture (response template generator and item selector) is trained in an end-to-end manner. 
Our approach combines the advantages of both classical slot filling approaches (that are generally controllable) and modern neural NLG approaches (that are generally more natural and accurate),
which brings both naturally sounded responses and more flexible item recommendation.

Another unique advantage of our NTRD lies in its zero-shot capability that can adapt with a regularly updated recommender system. 
Once a slotted response template is generated by the template generator,
different recommender systems could be plugged into the item selector easily to fill in the slots with proper items. 
Thus, NTRD can produce the diverse natural responses with the items recommended by different recommenders.

The contributions of this work are summarized as follows: 
(1) We present a novel framework called NTRD
for recommender dialogue system,
which decouples the response generation from the item recommendation via a two-stage strategy;
(2) Our NTRD first generates a response template that contains a mix of contextual words and slot locations explicitly associated with target items, and then fills in the slots with precise items by an item selector using a sufficient attention mechanism;
(3) Extensive experiments on standard dataset demonstrate our NTRD significantly outperforms previous state-of-the-art methods on both automatic metrics and human evaluation.
Besides, NTRD also exhibits the promising generalization ability on novel items that do not exist in training corpus.

\section{Related Work}

In this section, we first introduce the related work on task-oriented dialogue system. 
Then we review the existing literature on Conversational Recommender Systems (CRS), which can be roughly divided into two categories, \textit{i.e.}, attribute-centric CRS and open-ended CRS.

\cparagraph{Task-oriented Dialogue System.} 
\indent 
From the methodology perspective, there are two lines of the research on the task-oriented dialogue system, \textit{i.e.}, modular approaches~\cite{young2013pomdp} and end-to-end approaches~\cite{serban2016building,wen2016network,bordes2016learning,zhao2017generative,lei-etal-2018-sequicity}.
Recent works like GLMP~\cite{wu2019globaltolocal} and dynamic fusion network~\cite{qin2020dynamic} make attempt to dynamically incorporate the external knowledge bases into the end-to-end framework.
\citet{wu2019globaltolocal} introduce a global-to-local memory pointer network to RNN-based encoder-decoder framework to incorporate external knowledge in dialogue generation.
By contrast, our approach gets rid of pointer network paradigm and proposes a two-stage framework, which is modeled by the transformer-based architecture.

\cparagraph{Attribute-centric CRS.}
\indent
The attribute-centric CRS conducts the recommendations by asking clarification questions about the user preferences on a constrained set of item attributes. 
This kind of systems gradually narrow down the hypothesis space to search the optimal items according to the collected user preferences.
The various asking strategies have been extensively explored, such as memory network based approach~\cite{zhang2018towards}, entropy-ranking based approach~\cite{wu2018q20}, generalized binary search based approaches~\cite{zou2019learning,zou2020towards}, reinforcement learning based approaches~\cite{sun2018conversational,hu2018playing,chen2018learning,lei2020estimation,deng2021unified,li2021seamlessly}, adversarial learning based approach~\cite{ren2020crsal} and graph based approaches~\cite{xu2020user,lei2020interactive,ren2021learning,xu2021adapting}.
Most of these works~\citep{christakopoulou2018q,zhang2018towards,deng2021unified} retrieve questions/answers from a template pool and fill the pre-defined slots with optimal attributes.
Although this kind of systems are popular in the industry due to the easy implementation, 
they are still lack of the flexibility and the interactiveness, which leads to the undesirable user experience in practice.

\cparagraph{Open-ended CRS.}
\indent
Recently, researchers begin to explore the more free-style item recommendation in the response generation, \textit{i.e.}, open-ended CRS~\citep{li2018towards,chen2019towards,liao2019deep,kang2019recommendation,zhou2020improving,ma2020bridging,chen2020towards,liu2020towards,hayati2020inspired,zhou2020towards,zhang2021kecrs}.
Generally, this kind of systems consist of two major components, namely a recommender component to recommend items and a dialogue component to generate natural responses.
\citet{li2018towards} make the first attempt on this direction.
They release a benchmark dataset \textsc{ReDial} that collects human conversations about movie recommendation between paired crowd-workers with different roles (\textit{i.e.}, Seeker and Recommender).
Further studies~\citep{chen2019towards,zhou2020improving,ma2020bridging,sarkar2020suggest,lu2021revcore} leverage multiple external knowledge bases to enhance the performance of recommendation.
\citet{liu2020towards} propose a multi-goal driven conversation generation framework (MGCG) to proactively and naturally lead a conversation from a non-recommendation dialogue to a recommendation-oriented one.
Recently, \citet{zhou2021crslab} release an open-source CRS toolkit, \textit{i.e.}, CRSLab, to facilitate the research on this direction.
However, the Pointer Network~\citep{gulcehre2016pointing} or Copy Mechanism~\citep{gu2016incorporating} used in these approaches cannot be always accurately incorporated the recommended items into the generated replies.
Moreover, only the items mentioned in training corpus have a chance of being recommended in conversations by existing approaches.

Our work lies in the research of open-ended CRS. 
While in our work, we propose to decouple dialogue generation from the item recommendation.
Our approach first leverages Seq2Seq model~\citep{sutskever2014sequence} to generate the response template, and then fills the slots in template with proper items using sufficient multi-head self-attention mechanism.
Moreover, our work shows the unique advantage to produce novel items that do not exist in the training corpus.

\begin{figure*}[!t]
	\centering
	\includegraphics[width=1\textwidth]{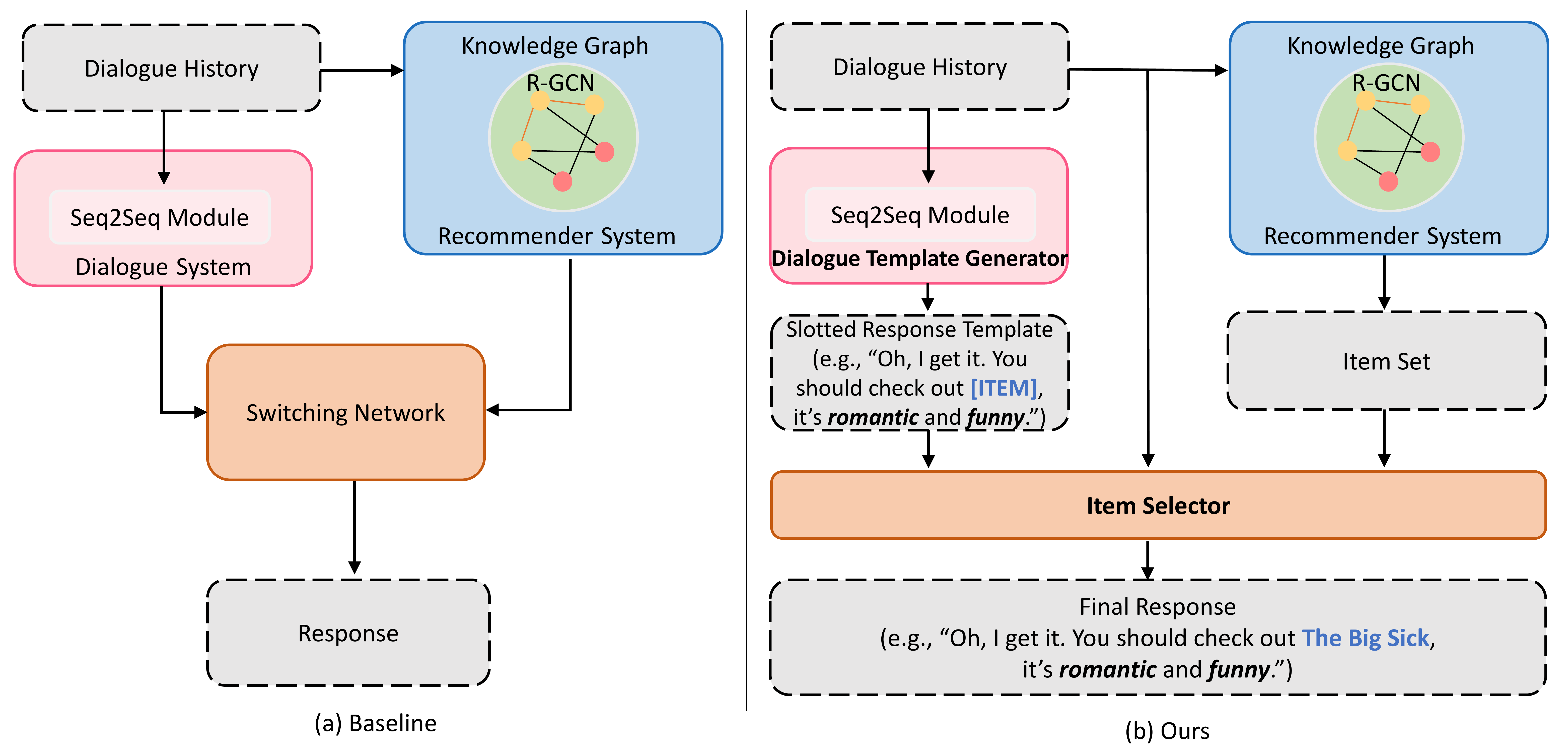}
	\caption{Comparison on modules of the existing frameworks and our proposed NTRD framework.}
	\label{fig:model_framework}
\end{figure*}

\section{Preliminary}

Formally, a dialogue consisting of $t$-turn conversation utterances is denoted as $\mathcal{D}=\left\{s_{t}\right\}_{t=1}^{N}$. 
Let $m$ denotes an item from the total item set $\mathcal{M}$, and $w$ denotes a word from vocabulary $\mathcal{V}$.
At the $t$-th turn, the recommender module chooses several candidate items $\mathcal{M}_{t}$ from the item set $\mathcal{M}$, 
while the dialogue module generates a natural language sentence $s_{t}$ containing a proper item $i$ from $\mathcal{M}_{t}$ to make recommendations. 
It is noteworthy that $\mathcal{M}_{t}$ can be equal to $\emptyset$ when there is no need for recommendation.
In that case, the dialogue module could continue to generate a chit-chat response or proactively explore the user's interests by asking questions.
To incorporate the recommended items into the generated reply, a switching mechanism~\citep{gulcehre2016pointing} or CopyNet~\citep{gu2016incorporating} is usually utilized to control the decoder to decide whether it should generate a word from the vocabulary or an item from the recommender output.
Specifically, the recommender predicts the probability distribution over the item set as $P_{\text {rec}}$,
and the dialogue module predicts the probability distribution over vocabulary as $P_{\text {dial}} \in \mathbb{R}^{|V|}$.
The overall probability of generating the next token is calculated as follows:
\begin{equation}
P(w_{o})=p_{s} P_{\text {dial}}(w)+\left(1-p_{s}\right) P_{\text {rec}}(i)
\end{equation}
\vspace{-1.5em}
\begin{equation}
p_{s}=\sigma\left(W_{s} e+b_{s}\right)
\end{equation}
where $w_{o}$ represents either a word from the vocabulary or an item from the item set, $e$ is the hidden representation in the final layer of the dialogue module.
$\sigma$ refers to the sigmoid function and $W_{s}$ and $b_{s}$ are the learnable parameters.

\section{Method}

In this section, we present the framework of learning \textbf{N}eural \textbf{T}emplates
for \textbf{R}ecommender \textbf{D}ialogue system, called \textbf{NTRD}.
As shown in Figure~\ref{fig:model_framework}, NTRD mainly consists of three components: a recommendation-aware response template generator, a context-aware item selector and a knowledge graph (KG) based recommender.
Given the dialogue context, the encoder-decoder based template generator focuses on generating the response template with item slots (Section~\ref{sec:temp_generation}).
Then the blank slots are filled by the item selector according to the dialogue context, candidate items from the recommender module and the generated response template (Section~\ref{sec:filler}).
Finally, the entire framework is trained in an end-to-end manner (Section~\ref{sec:objective}).

\begin{figure*}[h]
	\centering
	\includegraphics[width=0.7\textwidth]{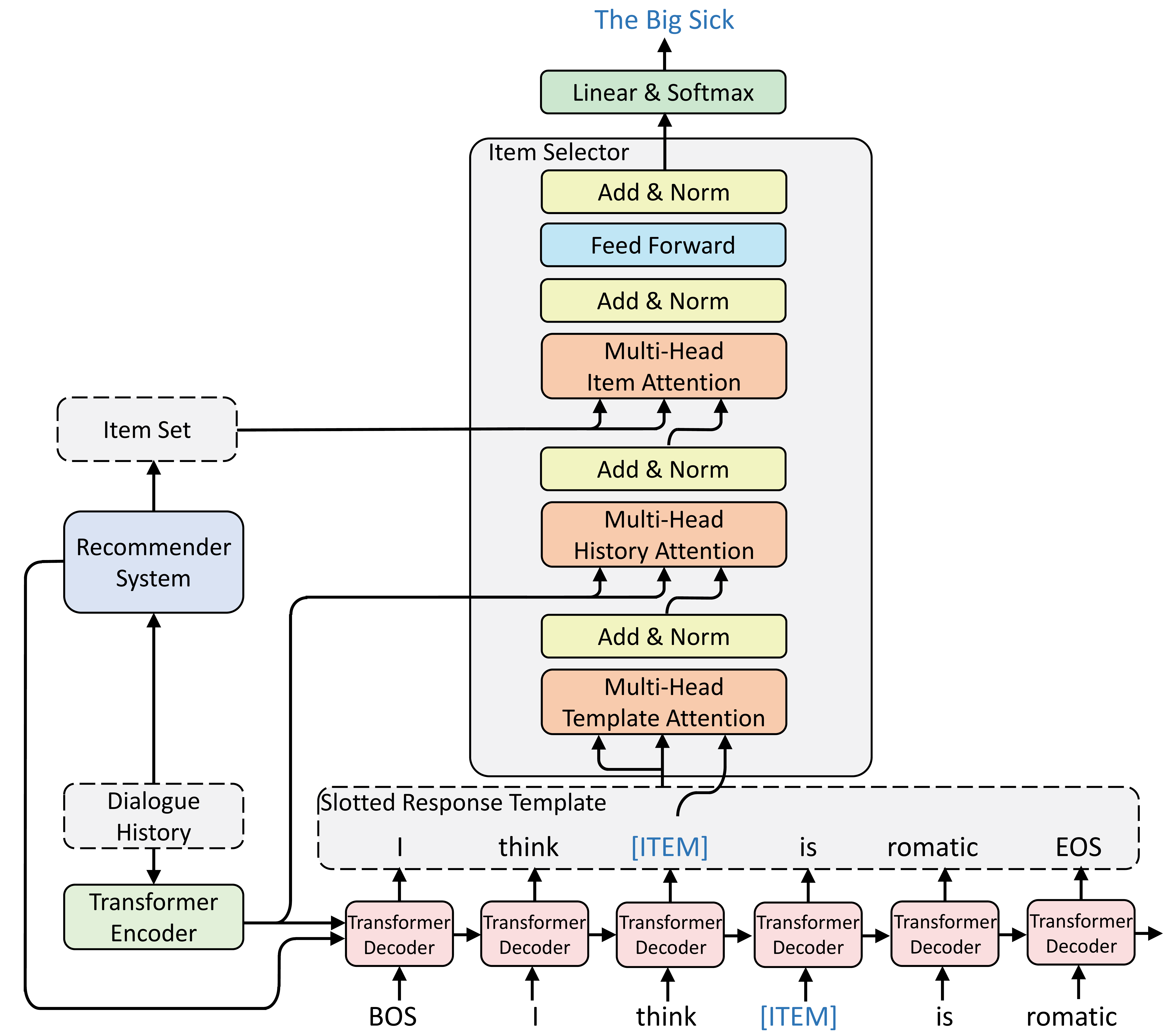}
	\caption{The overview of our approach. The slotted response template is first generated by the transformer decoder and then the item selector fills in the slots with proper items. Our framework enables sufficient information interaction among the generated template, dialogue history, and candidate items in a progressive manner, which is beneficial to selecting the more suitable items to fill in the slot locations.}
	\label{fig:model_overview}
\end{figure*}

\subsection{Response Template Generator}
\label{sec:temp_generation}

To generate the response template, we adopt the Transformer-based network~\citep{vaswani2017attention} to model the process.
Concretely, we follow~\citet{zhou2020improving} to use the standard Transformer encoder architecture and the KG-enhanced decoder which can effectively inject the information from KG into the generation process.
Then we add a special token \texttt{[ITEM]} into the vocabulary and mask all items in the utterances of dialogue corpus with \texttt{[ITEM]} tokens.
Thus, at each time step, the response template generator predicts either the special token \texttt{[ITEM]} or the general words from the vocabulary. 
Formally, the probability of generating the next token by the response template generator is given as follows:
\begin{equation}
P_{\text {dial}}(w)=softmax\left(W_{d} e+b_{d}\right)
\end{equation}
where $W_{d} \in \mathbb{R}^{|\mathcal{V}|\times d^{e}}$
and $b_{d} \in \mathbb{R}^{|\mathcal{V}|}$
are weight and bias parameters, $d^{e}$ is the embedding size of the hidden representation $e$.
After the generation process is finished, these special tokens serve as the item slots in generated templates, which will be filled with the specific items by item selector.

\subsection{Slot Filling with Item Selector}
\label{sec:filler}

Now we have the generated response templates, the rest we need to do is filling the slot locations with proper items.
Here, we first reuse the KG-enhanced recommender module~\citep{zhou2020improving} to get the user representation given the dialogue context. 
The recommender module learns a user representation $p_{u}$ through incorporating two special knowledge graphs, \textit{i.e.}, a word-oriented KG~\citep{speer2017conceptnet} to provide the relations between words and an item-oriented KG~\citep{bizer2009dbpedia} to provide the structured facts regarding the attributes of items.
Given the learned user preference $p_{u}$, we can compute the similarity between user and item as follows:
\begin{equation}
\label{eq:similarity}
similarity\left(m\right)=softmax\left(p_{u}^\mathrm{T}\cdot h_{m}\right)
\end{equation}
where $h_{m}$ is the learned embedding for item $m$, and $d^{h}$ is the dimension of $h_{m}$.
Hence, we rank all the items for $p_u$ according to Eq.~\ref{eq:similarity} and produce a candidate set from the total item set.

Existing works~\citep{chen2019towards,zhou2020improving} infer the final item only based on the dialogue context.
While the generated response template can also provide the additional information for selecting the final item.
For instance, as shown in the example of Figure~\ref{fig:model_framework}, the words \textit{\textbf{``romantic''}} and \textit{\textbf{``funny''}} after item slot could provide the contextual semantic information in the response for choosing the item to be recommended. 

Motivated by this, we propose a context-aware item selector by stacking sufficient multi-head attention blocks, as shown in Figure~\ref{fig:model_overview}. 
Formally, we define the embedding matrix $E_{slot}$ for all the slots in the template, where each slot embedding is the hidden representation from the final layer of transformer decoder. 
Similarly, the embedding matrix for the remaining tokens in the template is defined as $E_{word}$ and the embedding matrix output by the transformer encoder is $E_{ctx}$.
$H_{cand}$ is the concatenated embedding matrix for candidate items. Hence, the calculation in the item selector is conducted as follows:
\begin{equation}
\label{eq:selction_module}
\begin{aligned}
&\hat{E_{slot}} =\mathrm{MHA}\left(E_{slot}, E_{wrod}, E_{word}\right) \\
&E_{slot}^{\prime} =\operatorname{MHA}\left(\hat{E_{slot}}, E_{ctx}, E_{ctx}\right) \\
&E_{slot}^{\prime\prime} =\operatorname{MHA}\left(E_{slot}^{\prime}, H_{cand}, H_{cand}\right) 
\end{aligned}
\end{equation}
where $\operatorname{MHA}(Q, K, V)$ defines the multi-head attention function~\citep{vaswani2017attention} that takes a query matrix $Q$, a key matrix $K$, and a value matrix $V$
as input and outputs the attentive value matrix:
\begin{equation}
\begin{aligned}
&\operatorname{MHA}(Q,\!K,\!V)\!=\!\text {Concat\!}\left(\text {head}_{1}, \ldots, \text {head}_{\mathrm{h}}\right)\! W^{O} ,\\
&\text {where head }_{i}\!=\!\operatorname{Attention}\!\left(\!Q W_{i}^{Q}, K W_{i}^{K}, V W_{i}^{V}\!\right)
\end{aligned}
\end{equation}
Note that the layer normalization with residual connections and fully connected feed-forward network are omitted in Eq.~\ref{eq:selction_module} for simplicity.
By this means, the item selector is able to sufficiently fuse effective information from the generated template, dialogue context and candidate items in a progressive manner, which is beneficial to selecting the more suitable items to fill in the slot locations.

Finally, the item selector predicts a probability distribution over all items and selects the one with the highest score to fill in:
\begin{equation}
P_{\text {rec}}(w)=softmax\left(W_{r} e_{\text {slot}}+b_{r}\right)
\end{equation}
where $W_{r} \in \mathbb{R}^{|\mathcal{M}_{t}|\times d^{e}}$
and $b_{r} \in \mathbb{R}^{|\mathcal{M}_{t}|}$
are weight and bias parameters.

\subsection{Training Objectives}
\label{sec:objective}

Though the entire framework is typically two-stage, the two modules can be trained simultaneously in an end-to-end manner.
For the template generation, we optimize a standard cross-entropy loss as:
\begin{equation}
L_{\text {gen}}=-\sum_{t=1}^{N}  \log \left(P_{\text {dial}}(s_{t}|s_{1},...,s_{t-1})\right)
\end{equation}
where $N$ is the number of turns in a conversation $\mathcal{D}$, $s_{t}$ is the $t$-th utterance of the conversation. 

While the loss function for the item selector is calculated as:
\begin{equation}
L_{\text {slot}}=-\sum_{i=1}^{\mathcal{|M_{D}|}}  \log \left(P_{\text {rec}}(m_{i})\right)
\end{equation}
where $|\mathcal{M_{D}}|$ is the number of ground truth recommended items in a conversation $\mathcal{D}$.

We combine the template generation loss and the slot selecting loss as:
\begin{equation}
L=\lambda L_{\text {gen}} +  L_{\text {slot}}
\end{equation}
where $\lambda$ is a weighted hyperparameter.

During the inference, we apply greedy search to decoding the response template $s_{t}=(w_{1}, w_{2}, ..., w_{s})$. 
If $w_{i}$ is the special token \texttt{[ITEM]}, the item selector will be used to select the appropriate specific item based on the dialogue context, generated template and candidate items.
Finally, the completed response will be sent to the user to carry on the interaction.
\section{Experimental Setup}

\subsection{Dataset}

To evaluate the performance of our method, we conduct comprehensive experiments on the \textsc{ReDial} dataset\footnote{\url{https://redialdata.github.io/website/}}, which is a recent CRS benchmark~\citep{li2018towards}.
This dataset collects high-quality dialogues for recommendations on movies through crowd-sourcing workers on Amazon Mechanical Turk (AMT).
It contains 10,006 conversations consisting of 182,150 utterances related to 6,924 movies, which is split into the training, validation, and test set in an 80-10-10 proportion.
\subsection{Evaluation Metrics}

Both automatic metrics and human evaluation are employed to evaluate the performance of our method. 
For dialogue generation, 
automatic metrics include:
(1) \textbf{Fluency}: perplexity (PPL) measures the confidence of the
generated responses.
(2) \textbf{Diversity}: Distinct-n (Dist-n)~\citep{li2016diversity} are defined as the number of distinct n-grams divided by the total amount of words. Specifically, we use Dist-2/3/4 at the sentence level to evaluate the diversity of generated responses. 

For recommendation task, existing works~\citep{li2018towards,chen2019towards,zhou2020improving} individually evaluate the performance on recommendation using Recall@k.
However, the goal of open-ended CRS is to smoothly chat with users and naturally incorporate proper recommendation items into the responses.
In other words, it is important for the system to generate informative replies containing the accurate items.
Hence, we introduce a new metric that checks whether the ground-truth item is included in the final generated response, \textit{i.e.}, Recall@1 in Response (ReR@1).
Similarly, if the generated response has an item token, we calculate whether the top-k (k=10, k=50) items of the probability distribution for this position contain the ground truth item, \textit{i.e.}, ReR@10 and ReR@50. 
Besides, we also introduce the Item Diversity that measures the percentage of the recommended items mentioned in the generated response to all items in the dataset.
Item Ratio is introduced by \citet{zhou2020improving} to measure the ratio of items in the generated response.

For human evaluation, 100 dialogues are randomly sampled from the test set. 
Then three crowd-workers are employed to score on the generated responses in terms of \textit{Fluency} and \textit{Informativeness}.
The range of score is 1 to 3.
The higher score means the better.
The average score of each metric on these 100 dialogues evaluated by three workers is reported.
The inter-annotator agreement is measured by Fleiss’ Kappa~\citep{fleiss1973equivalence}.

\subsection{Implementation Details}

The models are implemented in PyTorch and trained on one NVIDIA Tesla V100 32G card.
For the fair comparison, we keep the data pre-processing steps and hyperparameter settings the same as the KGSF model~\citep{zhou2020improving} in the released implementation\footnote{\url{https://github.com/RUCAIBox/KGSF}}.
The embedding size $d^{h}$ of the item in recommender module is set to 128, and the embedding size $d^{e}$ in dialogue module is set to 300.
We follow the procedure in KGSF to pre-train the knowledge graph in the recommender module using Mutual Information Maximization (MIM) loss for 3 epochs. 
Then the recommender module is trained until the cross-entropy loss converges.
For the training of response template generator, we replace the movies mentioned in the corpus with a special token \texttt{[ITEM]} and add it to the vocabulary. 
We use Adam optimizer with the $1e-3$ learning rate. The batch size is set to 32 and gradient clipping restricts in [0, 0.1].
The generation loss and the item selection loss are trained simultaneously with the weight $\lambda=5$.

\subsection{Baselines}
We introduce the baseline models for the experiments in the following:
\begin{itemize}

\item \textbf{\textsc{ReDial}}~\citep{li2018towards}: The baseline model proposed by ~\citet{li2018towards} consists of an auto-encoder~\citep{wang2015collaborative} recommender, a dialogue generation model based on HRED~\citep{serban2017hierarchical} and a sentiment prediction model.

\item \textbf{KBRD}~\citep{chen2019towards}: 
This model utilizes a KG to enhance the user representation. The transformer-based~\citep{vaswani2017attention} dialogue generation model uses KG information as the vocabulary bias for generation.

\item \textbf{KGSF}~\citep{zhou2020improving}: The model proposes to incorporate two external knowledge graphs, \textit{i.e.}, a word-oriented KG and an item-oriented KG, to further enhance in modeling the user preferences.

\end{itemize}

\begin{table*}[htbp]
\centering
\resizebox{1\textwidth}{!}{
\begin{tabular}{lccccccccc}
\hline
Model  & PPL           & Dist-2         & Dist-3        & Dist-4         & ReR@1     & ReR@10      & ReR@50     & Item Diversity & Item Ratio \\ \hline
\textsc{ReDial}~\citep{li2018towards} & 28.1          & 0.225          & 0.236         & 0.228          & -              & -               & -               & -              & 15.8           \\
KBRD~\citep{chen2019towards}   & 17.9          & 0.263          & 0.368         & 0.423          & -              & -               & -               & -              & 29.6           \\
KGSF~\citep{zhou2020improving}   & 5.55          & 0.305          & 0.466         & 0.589          & 0.889          & 1.083           &  1.733           & 6.03           & 31.5           \\
\textbf{NTRD (ours)}   & \textbf{4.41} & \textbf{0.578} & \textbf{0.820} & \textbf{1.005} & \textbf{1.806} & \textbf{12.503} & \textbf{31.592} & \textbf{11.05} & \textbf{66.77} \\ \hline
\end{tabular}
}
\caption{\label{tab:sota_comparison}
Automatic evaluation results on the \textsc{ReDial} dataset. Numbers in bold denote that the improvement over the best performing baseline is statistically significant. }
\end{table*}

\begin{table}[htbp]
\resizebox{77mm}{!}{
\begin{tabular}{lccc}
\hline
Method & Fluency & Informativeness & Kappa \\ \hline
KGSF  &    2.24    &        1.92        &   0.67   \\
\textbf{NTRD (ours)}  &   \textbf{2.48}      &       \textbf{2.16}          &  0.62     \\ \hline
Human  &    2.85    &        2.37        &   0.74   \\ \hline
\end{tabular}
}
\caption{\label{tab:human_eval} Human evaluation results on the dialogue generation. Human
responses come from the ground truth responses provided in the dataset. 
}
\end{table}

\section{Experimental Results}

\subsection{Evaluation on Dialogue Generation}

We conduct the automatic and human evaluations to evaluate the quality of generated responses.

\cparagraph{Automatic Evaluation.}
\indent
Table~\ref{tab:sota_comparison} shows the automatic evaluation results of the baseline models and our proposed NTRD on dialogue generation.
As we can see, our NTRD is obviously better on all automatic metrics compared to the baseline models.
Specifically, NTRD achieves the best performance on PPL, which indicates the generator of NTRD can also generate the fluent response templates.
In terms of diversity, NTRD consistently outperforms the baselines with a large margin on Dist-2/3/4.
This is because the generated template provides the extra contextual information for slot filling so as to produce more diverse and informative responses.

\cparagraph{Human Evaluation.}
\indent
We report the human evaluation results in Table~\ref{tab:human_eval}.
All Fleiss’s kappa values exceed 0.6, indicating crowd-sourcing annotators have reached the substantial agreement.
Compared to KGSF, our NTRD performs better in terms of Fluency and Informativeness. 
NTRD decouples the response generation and item injection by first learning response templates and then filling the slots with proper items.
Hence, it can generate the more fluent and informative responses in practice.

\begin{table}[ht]
\centering
\begin{tabular}{lc}
\hline
Model & Novel Ratio  \\ \hline
KGSF           & 0.00                                   \\
\textbf{NTRD (ours)}           & \textbf{13.40}                     \\ \hline
\end{tabular}
\caption{\label{tab:novel_exp} Evaluation results of the experiments on the ability of predicting the novel item.}
\end{table}

\begin{table}[ht]
\small
\centering
\begin{tabular}{rp{6cm}}
\toprule
\multicolumn{2}{c}{{\bf Conversation Case}} \\
\midrule
\textbf{User}: & Hello! \\
\midrule
\textbf{Human}: & What kind of movies do you like? \\
\textbf{KGSF}: & Hi! What kind of movies do you like? \\
\textbf{NTRD}: & Hello, what type of movies are you looking for? \\
\midrule
\textbf{User}: & I love \textcolor{red}{comedy} movies like \textit{\textcolor{blue}{Blades of Glory}}. \\
\midrule
\textbf{Human}: & I see, you can watch \textit{\textcolor{blue}{Shaun of the Dead}}. \\
\textbf{KGSF}: & I would recommend \textit{\textcolor{blue}{Bad Santa}}. \\
\textbf{NTRD}: & How about \textit{\textcolor{blue}{Anchorman 2: The Legend Continues}}. \\
\midrule
\textbf{User}: & I saw that one. A \textcolor{red}{classic parody} film. \\
\midrule
\textbf{Human}: & Also \textit{\textit{\textcolor{blue}{Hot\_Fuzz}}} is funny. I really though \textit{\textcolor{blue}{Ghostbusters}} was funny. Or you can watch \textit{\textcolor{blue}{Hangover}}. \\
\textbf{KGSF}: & I would recommend \textit{\textit{\textcolor{blue}{Bad Santa}}}. \\
\textbf{NTRD}: & Have you seen \textit{\textcolor{blue}{Shaun of the Dead}}? It's a really good one. \\
\midrule
\textbf{User}: & I am a big \textcolor{red}{fan} of both of those movies as well. \\
\midrule
\textbf{Human}: & But \textit{\textcolor{blue}{Deadpool 2 (2018)}} just came out and it 's funny as well. Have you seen it? \\
\textbf{KGSF}: & I also like \textit{\textcolor{blue}{Hot Fuzz}}. \\
\textbf{NTRD}: & I think you will like \textit{\textcolor{blue}{Borat}}. \\
\midrule
\textbf{User}: & Not yet, but I plan too. \\
\midrule
\textbf{Human}: & Ok. Hope I helped, bye. \\
\textbf{KGSF}: & I also like \textit{\textcolor{blue}{Groundhog Day}}. \\
\textbf{NTRD}: & I hope you enjoy your night! \\
\bottomrule
\end{tabular}
\caption{\label{tab:case_study} A sampled case between a real user as a seeker, and the dialogue agents (including human, KGSF and our NTRD) as recommenders.
Items mentioned are marked in the \textcolor{blue}{blue} color, while the user preferences in user’s turn are marked in the \textcolor{red}{red} color.}
\end{table}

\subsection{Evaluation on Recommendation}

In this section, we evaluate the performance of recommendation from two aspects, \textit{i.e.}, conversational item recommendation to assess the recall performance and novel item recommendation to investigate the generalization ability.

\cparagraph{Conventional Item Recommendation.}
\indent
To further investigate the performance of NTRD on the conventional item recommendation, we present the experimental results of ReR@k (k=1, 10 and 50), Item Diversity and Item Ratio in Table~\ref{tab:sota_comparison}.
As can be seen, when evaluating the actual performance of recommendation based on final produced responses, the state-of-the-art method KGSF performs poorly with only 0.889\% ReR@1.
This indicates the switching network in KGSF cannot accurately incorporate the recalled items into the generated responses.
It violates the original intention of the open-ended CRS, \textit{i.e.}, to not only smoothly chat with users but also recommend precise items using free-form natural text.
By contrast, our NTRD framework performs significantly better, which shows the decoupling strategy brings an obvious advantage of incorporating the precise items into the conversations with users.
Furthermore, NTRD achieves the highest item ratio and item diversity.
On the one hand, the template generator introduces a special token \texttt{[ITEM]} and thus reduces the size of vocabulary, which would increase the predicted probability of item slot during the generation process. 
On the other hand, the item selector utilizes sufficient information from dialogue context, generated template and candidate items to help select the high-quality recommended items.

\cparagraph{Novel Item Recommendation.}
\indent
Existing methods have one major drawback that they cannot handle the novel items never appearing in the training corpus.
To validate the unique advantage of our NTRD on novel item recommendation, we conduct an additional experiment.
Specifically, we collect all items from the test set that do not appear in the training set, \textit{i.e.}, 373 novel items in total.
To learn the representations of these novel items, we first include them together with other ordinary items in the pre-training of the recommender modules of both KGSF and NTRD.
However, when training the dialogue modules, we only use the normal training set where these novel items are excluded.
Then we evaluate the models on the test set. 
As we can see in Table~\ref{tab:novel_exp}, the 13.40\% (50 of 373) of novel items can be successfully incorporated into the final responses and thus recommended to the users, while KGSF fails to recommend any of the novel items.
This verifies the promising generalization ability of NTRD on the unseen items, which is crucial in the real-world settings.
Even though the recommender modules would update regularly to include novel items, our framework can still effectively incorporate these items into conversational responses without re-training the dialogue model.

\subsection{Case Study}

In this section, we present an interactive example to demonstrate how our model works in the interactive dialogue. More cases please refer to the Appendix.
In Table~\ref{tab:case_study}, we mark all the mentioned items in blue color and the user preferences in red color for easy reading.
The outputs of three recommenders including human, KGSF and NTRD are presented for the comparison.
The dialogue starts with greetings between the user (seeker) and the recommenders, followed by the recommenders proactively seeking user preference by asking which kind of movie he/she likes. 
With the focused preference of the user on ``\textit{comedy}'' movies, the recommenders provide some candidate movie items with the interesting comments.
The responses of KGSF tend to be boring and dull, and it does not switch its recommendation item even though the user reject it by saying ``I saw that one''.
Instead, our NTRD behaves more like human and shares a similar taste with the human recommender (both recommend the comedy movie ``\textit{Shaun of the Dead}'').
Then when the user expresses intention to watch, both our NTRD and human recommender respond with goodwill and end the dialogue, while KGSF still tries to give a new movie recommendation.
In a nutshell, our NTRD is able to provide informative and interesting item recommendation at the utterance level, and behaves more naturally and interactively at the dialogue level.

\section{Conclusions}
In this paper, we introduce \textbf{NTRD}, a novel recommender dialogue framework that decouples the dialogue generation from the item recommendation via a two-stage strategy. 
Our approach reconciles classical slot filling approaches with modern neural NLG approaches, which make the recommender dialogue system more flexible and controllable.
Besides, our approach exhibits the promising generalization ability to recommend novel items that do not appear in the training corpus. 
Extensive experiments show our approach significantly outperforms the previous state-of-the-art methods.

For future work, the generalization ability of NTRD could be further explored.
Current method only supports one placeholder with broad semantics to represent all the item mentions in the dialogue corpus, which is lack of fine-grained annotation.
One possible attempt is to extend it to support fine-grained item placeholders, such as replacing the placeholder with different attributes of the items, to further improve its performance.


\bibliography{anthology,custom}
\bibliographystyle{acl_natbib}

\appendix

\section{Appendix}
\label{sec:case_appendix}

In this section, we show more conversation examples. 
We mark all the mentioned items in blue
color and the user preferences in red color for easy
reading.

\begin{table}[ht]
\small
\centering
\begin{tabular}{rp{6cm}}
\toprule
\multicolumn{2}{c}{{\bf Conversation Case}} \\
\midrule
\textbf{User}: & Hi. \\
\midrule
\textbf{Human}: & Tell me what kind of movies would you like to watch? \\
\textbf{KGSF}: & What kind of movies do you like? \\
\textbf{NTRD}: & What type of movies do you like? \\
\midrule
\textbf{User}: & I like \textcolor{red}{horror} and \textcolor{red}{thriller}. Especially low budget like \textit{\textcolor{blue}{The Boy (2016 film)}} and \textit{\textcolor{blue}{Don't Breathe}}.  \\
\midrule
\textbf{Human}: & Have you seen \textit{\textcolor{blue}{The Babadook}}? \\
\textbf{KGSF}: & I haven't seen either. I heard of that one. I will be \textit{\textcolor{blue}{Don't Breathe}} and \textit{\textcolor{blue}{Insidious: Chapter 2}} and \textit{\textcolor{blue}{Insidious: Chapter 2}}. \\
\textbf{NTRD}: & Have you seen \textit{\textcolor{blue}{Sweeney Todd: The Demon Barber of Fleet Street}}? \\
\midrule
\textbf{User}: & No, but I want to! \\
\midrule
\textbf{Human}: & What about \textit{\textcolor{blue}{The Cabin in the Woods}}? \\
\textbf{KGSF}: & I haven't seen either. I heard of that one. I will be \textit{\textcolor{blue}{Don't Breathe}}. \\
\textbf{NTRD}: & \textit{\textcolor{blue}{Sweeney Todd: The Demon Barber of Fleet Street}} is a great one. \\
\midrule
\textbf{User}: & Yes, I like that one. \\
\midrule
\textbf{Human}: & I really enjoy \textit{\textcolor{blue}{The Cabin in the Woods}}? \\
\textbf{KGSF}: & I haven't seen either. I heard of that one. I will be \textit{\textcolor{blue}{Don't Breathe}}. \\
\textbf{NTRD}: & Yes, \textit{\textcolor{blue}{It (2017 film)}} is really good. \\
\midrule
\textbf{User}: & Yeah, me too. \\
\midrule
\textbf{Human}: & Have you seen \textit{\textcolor{blue}{Raw (film)}}? It is pretty scary. \\
\textbf{KGSF}: & I hope you enjoy them! \\
\textbf{NTRD}: & Yes, \textit{\textcolor{blue}{Don't Breathe}} is really good. \\
\midrule
\textbf{User}: & No, I haven't heard of that. Thanks for the suggestions. \\
\midrule
\textbf{Human}: & Bye. You are Welcome. \\
\textbf{KGSF}: & You 're welcome. Bye. \\
\textbf{NTRD}: & You are welcome. Enjoy your day. \\
\bottomrule
\end{tabular}
\caption{\label{tab:case_append_1} Example (A). }
\end{table}

\begin{table}[h]
\small
\centering
\begin{tabular}{rp{6cm}}
\toprule
\multicolumn{2}{c}{{\bf Conversation Case}} \\
\midrule
\textbf{User}: & User: I like to watch \textcolor{red}{scary} movies. \\
\midrule
\textbf{Human}: & Tell me what kind of movie do you like ? Did you watch \textit{\textcolor{blue}{It (2017 film)}}? And \textit{\textcolor{blue}{It (1990 film)}}? \\
\textbf{KGSF}: & I would recommend some scary ones, are you like \textit{\textcolor{blue}{Scream (1996 film)}} or \textit{\textcolor{blue}{Annabelle}} and \textit{\textcolor{blue}{Insidious: Chapter 2}}? \\
\textbf{NTRD}: & What about \textit{\textcolor{blue}{The Shining}}? \\
\midrule
\textbf{User}: & I have seen the old one.         \\
\midrule
\textbf{Human}: & Ok, you should watch the new one. Even though i would rather the old one. And did you watch \textit{\textcolor{blue}{The Last Exorcism}}? I hope i have helped. \\
\textbf{KGSF}: & I would recommend   \textit{\textcolor{blue}{The Babadook}}. \\
\textbf{NTRD}: & Have you seen \textit{\textcolor{blue}{It Follows}}? That 's a great one and is a great one of my favorites. \\
\midrule
\textbf{User}: & No, I have not but it sounds \textcolor{red}{scary}! Thank you for the suggestions!   \\
\midrule
\textbf{Human}: & You should try them. You 're welcome. Good bye. \\
\textbf{KGSF}: & Bye. \\
\textbf{NTRD}: & You 're welcome. Bye. \\
\bottomrule
\end{tabular}
\caption{\label{tab:case_append_1} Example (B). }
\end{table}

\begin{table}[ht]
\small
\centering
\begin{tabular}{rp{6cm}}
\toprule
\multicolumn{2}{c}{{\bf Conversation Case}} \\
\midrule
\textbf{User}: & Hello , are you a movie buff ? \\
\midrule
\textbf{Human}: & Hi, I guess I could say yes from the ridiculous stack of movies :) You? \\
\textbf{KGSF}: & Hello! What kind of movies do you like? \\
\textbf{NTRD}: & Hi, what type of movies do you like? \\
\midrule
\textbf{User}: & I watch many movies. I especially like \textcolor{red}{science fiction}. Any recommendations ?       \\
\midrule
\textbf{Human}: & An oldy but a goody is \textit{\textcolor{blue}{The Wizard of Oz (1939)}}. I prefer more actions and comedies, any recomendations? \\
\textbf{KGSF}: & I haven't seen that one. I 'm glad I could help. \\
\textbf{NTRD}: & How about \textit{\textcolor{blue}{Alien: Covenant (2017)}}? \\
\midrule
\textbf{User}: & I like \textcolor{red}{Bruce Willis action}.  \\
\midrule
\textbf{Human}: & I really enjoy movies like \textit{\textcolor{blue}{The Avengers  (2012)}} and \textit{\textcolor{blue}{Iron Man  (2008)}} . Recently we watched \textit{\textcolor{blue}{Spider-Man: Homecoming (2017)}} and it was great! \\
\textbf{KGSF}: & I haven't seen that one. I will be good. I will be good. I will be \\
\textbf{NTRD}: & \textit{\textcolor{blue}{Casablanca}} is good as well. \\
\midrule
\textbf{User}: & I like \textit{\textcolor{blue}{Seven Samori}} as well as \textit{\textcolor{blue}{The Magnificent Seven}}.   \\
\midrule
\textbf{Human}: & I like Bruce Willis in a lot of movies, especially in \textit{\textcolor{blue}{Die Hard (film series)}}. \\
\textbf{KGSF}: & I haven't seen that one. I will be good. I will be good. I will be \\
\textbf{NTRD}: & \textit{\textcolor{blue}{Iron Man (2008)}} was good too. \\
\midrule
\textbf{User}: & I will watch \textit{\textcolor{blue}{The Avengers(2012)}} again , Thanks!  \\
\midrule
\textbf{Human}: & I will add it to the Netflix que. Thank you! \\
\textbf{KGSF}: & Have a great day! \\
\textbf{NTRD}: & You too. \\
\bottomrule
\end{tabular}
\caption{\label{tab:case_append_1} Example (C).}
\end{table}




\end{document}